\documentclass[journal]{IEEEtran}
\usepackage{latexsym,,amssymb,amsmath,graphicx,epsf,cite,bbm,float}
\usepackage{ifpdf}
\usepackage{epstopdf}

\usepackage{algorithm,algorithmic}
\usepackage{amsmath,amssymb,bm}
\usepackage{amsfonts,dsfont,color,bbm,subcaption}

\newcommand{\abs}[1]{|#1|}
\newcommand{\bs}[1]{\boldsymbol{#1}}

\DeclareMathOperator{\sign}{sgn}

\usepackage{hyperref,amsthm}

\newtheorem{lemma}[]{Lemma}

\newtheorem{proposition}[]{Proposition}

\newtheorem{remark}[]{Remark}

\newtheorem{definition}[]{Definition}

\newtheorem{assumption}[]{Assumption}

\begin{document}

\title{A Note On Nonlinear Regression Under L2 Loss} 
\author{\IEEEauthorblockN{Kaan Gokcesu}, \IEEEauthorblockN{Hakan Gokcesu} }
\maketitle

\begin{abstract}
	We investigate the nonlinear regression problem under L2 loss (square loss) functions. Traditional nonlinear regression models often result in non-convex optimization problems with respect to the parameter set. We show that a convex nonlinear regression model exists for the traditional least squares problem, which can be a promising towards designing more complex systems with easier to train models.
\end{abstract}

\section{Introduction}


Nonlinear regression is a statistical technique used to model the relationship between a target variable and one or more feature variables \cite{neter1996applied,weisberg2005applied}. It is an extension of the linear regression, which assumes that the relationship between the target and the feature variables is linear. Nonlinear regression models are widely used in many fields, including but not limited to; biology, engineering, finance, economics, physics and agriculture \cite{myers2016response,bates1988nonlinear,yagiz2010application,bulturbayevich2022application,drignei2008parameter,archontoulis2015nonlinear,tiedeman1998application}.

Unlike linear regression models, nonlinear regression models cannot be solved analytically and require numerical methods to estimate its parameters \cite{weisberg2005applied}.
Nonlinear regression models are powerful in that they can be used to model a wide variety of functional forms such as exponential, logarithmic, power, and polynomial functions \cite{seber2003nonlinear}. These models may allow for more flexible and nuanced analyses of complex relationships between input variables, which makes it a valuable tool for scientific research and data analysis.

One of the earliest examples of nonlinear regression can be traced back to the work of Francis Galton in the late 19th century \cite{galton1886regression}. Galton used nonlinear regression to study the relationship between the heights of parents and their offspring, where the data was modeled with a quadratic function \cite{galton1886regression}. Since then, nonlinear regression has been refined and expanded upon by numerous researchers. In recent years, advances in computing power and statistical software have made nonlinear regression more accessible to researchers. However, caution is still required in the interpretation of the results, as these models can be more sensitive to outliers and model assumptions than linear models (presumably because of over-fitting) \cite{bates1988nonlinear}.

One common application of nonlinear regression is in pharmacokinetic modeling, which involves studying the absorption, distribution, metabolism, and excretion of drugs in the body \cite{bates1988nonlinear}. Nonlinear regression models are used to estimate the pharmacokinetic parameters of a drug such as its clearance rate and volume of distribution, which are important for determining the appropriate dosage and administration schedule.
Another example of nonlinear regression is in the field of ecology, where it is used to model population dynamics and species interactions \cite{weisberg2005applied}. Nonlinear models can account for complex ecological processes, such as density-dependent population growth and predator-prey interactions, which are not easily captured by linear models.

In many machine learning applications, the goal is generally to train a model that makes predictions based on the input data \cite{alpaydin2020introduction}. In such problems, nonlinear regression models can be fit using a variety of methods, including maximum likelihood estimation, least squares regression, and Bayesian methods \cite{bates1988nonlinear}. While the choice of method depends on the specific application and the characteristics of the data; typically a loss function that measures how well the model is performing on a given task is utilized.
One such common loss is the traditional square (L2) loss function \cite{bache2013uci}, which measures the squared difference between the predicted values of a model and the actual values \cite{goodfellow2016deep,zhu2009introduction,huber1973robust,gokcesu2021generalized,gokcesu2022nonconvex,gokcesu2021optimal}. It is also dubbed as the mean squared error (MSE) analysis \cite{bishop2006pattern}. It is commonly used in many applications, especially when the data distribution is normal and the outliers are not dominant \cite{brownlee2018deep}. For this loss, while linear models have analytical solutions, nonlinear models do not.

Henceforth, one important consideration in nonlinear regression is the choice of optimization method used to estimate the parameters of the model. 
One of the advantages of using the L2 loss function is that it is differentiable, which means that it can be used in gradient-based optimization algorithms like stochastic gradient descent (SGD). This allows for sequentially updating the model parameters in a way that minimizes the loss function, which in turn improves the model's performance on the task at hand \cite{rumelhart1986learning}. The L2 loss is suitable for not only for measuring the performance of a model but also for optimizing its parameters for improved performance \cite{hastie2009elements}. 

Under squared loss, gradient-based methods, such as the Gauss-Newton method or the Levenberg-Marquardt algorithm, can be utilized in nonlinear regression. These methods iteratively adjust the parameters of the model until they converge on the values that best fit the data \cite{gill2019practical,gokcesu2021nonparametric}.
Due to nonlinear regression capability for modeling complex relationships between variables \cite{weisberg2005applied}, it has wide applicability and flexibility. By using nonlinear regression, the underlying mechanism that drive the data can be better understood, which can lead to more accurate predictions and better-informed decisions \cite{liao1994interpreting}. 

Model assumptions and optimization methods needs to be carefully considered to ensure the accuracy and reliability of the results. There exist various ways to introduce nonlinearity into the model such as isotonic regression \cite{gokcesu2021efficient,barlow1972isotonic}. More commonly, nonlinearity is introduced via transform mappings or activation functions \cite{brownlee2018deep,hastie2009elements}. When such mappings are utilized, the objective generally becomes a nonconvex optimization, which has much slower convergence \cite{gokcesu2022low,gokcesu2022efficient}. 

We show that, under the square loss objective, there exist nonlinear models that preserve convexity. We first provide some preliminaries and show that for arbitrary convex losses only linear models preserve convexity. Then, we show that specifically for the L2 loss functions, there exist nonlinear models with convex objectives.

\section{Preliminaries}
We first provide the standard problem setting, where we have the target variables $\{y_n\}_{n=1}^N$ and the features $\{\bs{x_n}\}_{n=1}^N$. Let us have the linear weights $\bs{w}$, and a possibly nonlinear transform $g(\cdot)$ to create our estimates, i.e.,
\begin{align}
	\hat{y}_n=&g(\bs{w^Tx_n}),&&\forall n
\end{align} 
Let us have the general cumulative loss
\begin{align}
	L(\bs{w})=\sum_{n=1}^{N}l(g(\bs{w^Tx_n}),y_n).
\end{align}
Since convexity is preserved under summation; if each individual loss $l(g(\bs{w^Tx_n}),y_n)$ is convex in $\bs{w}$; the cumulative loss $L(\bs{w})$ will also be convex in $\bs{w}$.

To analyze the convexity of $l(g(\bs{w^Tx}),y)$ for some $\bs{x}$ and $y$; we can look at its Hessian. The respective gradient and Hessian are as follows where $z(\bs{w})=\bs{w^Tx}$:
\begin{align}
	\nabla_{\bs{w}}l(g(\bs{w^Tx}),y)=&\dfrac{dl}{dg}(g(z(\bs{w})),y)\dfrac{dg}{dz}(\bs{w^Tx})\bs{x},\\
	\nabla^2_{\bs{w}}l(g(\bs{w^Tx}),y)=&\dfrac{d^2l}{dg^2}(g(z(\bs{w})),y)\left(\dfrac{dg}{dz}(z(\bs{w}))\right)^2\bs{x}\bs{x^T}\nonumber
	\\&+\dfrac{dl}{dg}(g(z(\bs{w})),y)\dfrac{d^2g}{dz^2}(z(\bs{w}))\bs{x}\bs{x^T},
\end{align}
where the derivatives are with respect to their scalar arguments.
To guarantee positive semi-definiteness, i.e., 
\begin{align}
	\nabla^2_{\bs{w}}l(g(\bs{w^Tx}),y)\succeq0
\end{align}
we need the following with some abuse of notation:
\begin{align}
	l''(g(\bs{w^Tx}),y)g'^2(\bs{w^Tx})+l'(g(\bs{w^Tx}),y)g''(\bs{w^Tx})\geq0.
\end{align}

For arbitrary convex losses $l(\cdot,y)$, this inequality is only guaranteed for when $g''(\bs{w^Tx})=0$, i.e., only the linear model $g(\bs{w^Tx})=\alpha\bs{w^Tx}+\beta$. Hence, for arbitrary convex functions, the nonlinearity introduced by $g(\cdot)$ will invalidate the convexity property of the loss function. However, given a specific loss function, we may be able to design an appropriate nonlinearity. First, we have the following definition.
\begin{definition}
	Let $z=\bs{w^Tx}$. We define the following equality
	\begin{align*}
		l_y(g(z))=l(g(\bs{w^Tx}),y).
	\end{align*}
\end{definition}
Secondly, since $g(\cdot)$ may not necessarily have continuous second derivative. To this end, we have the following property.
\begin{proposition}
	If $l_y(g(z))$ is convex in $z$, than $l(g(\bs{w^Tx}),y)$ is convex in $\bs{w}$. 
	\begin{proof}
		To see this, let $\bs{w}=\lambda\bs{w_1}+(1-\lambda)\bs{w_2}$ and consequently $z=\lambda z_1+(1-\lambda)z_2$. We have
		\begin{align}
			l(g(\bs{w^Tx}),y)=&l(g(\lambda\bs{w_1^Tx}+(1-\lambda)\bs{w_2^Tx}),y),
			\\=&l(g(\lambda z_1+(1-\lambda)z_2),y),
			\\=&l_y(g(\lambda z_1+(1-\lambda)z_2)),
			\\\leq&\lambda l_y(g(z_1)+(1-\lambda)l_y(g(z_2)),
			\\\leq&\lambda l(g(\bs{w_1^Tx}),y)+(1-\lambda)l(g(\bs{w_2^Tx}),y),
		\end{align}
		which concludes the proof.
	\end{proof}
\end{proposition}

\section{Convex Nonlinear Model for Square Loss}
Our goal is to find a nonlinear $g(\cdot)$ such that $l_y(g(z))$ is convex in $z$. We focus on a specific type of loss function, which is the L2 loss (the squared loss), i.e., 
\begin{align}
	l(g(\bs{w^Tx}),y)=l_y(g(z))=(g(z)-y)^2.
\end{align}
\begin{assumption}\label{ass:diff}
	$g(\cdot)$ is differentiable with a continuous first derivative $g'(\cdot)$.
\end{assumption}
\begin{remark}
	If \autoref{ass:diff} holds, the convexity of $l_{y}(g(z))$ in $z$ is directly implied if its first derivative is nondecreasing with $z$.
\end{remark}
\noindent The first derivative of $l_y(g(z))$ is given by
\begin{align}
	\dfrac{dl_{y}(g(z))}{dz}=l_{y}'(g(z))=&2(g(z)-y)g'(z),\label{eq:dl}
\end{align}
from the chain rule when \autoref{ass:diff} holds.

Ideally, we want the nonlinear transform $g(\cdot)$ to be a one-to-one mapping. To this end, we have the following assumption.
\begin{assumption}
	The first derivative of $g(\cdot)$ is always positive, i.e., $g'(z)>0$ for all $z$. 
\end{assumption}
\noindent By design, we investigate the class of functions that are odd symmetric, i.e.,
\begin{align}
	g(z)=-g(-z),
\end{align}
and consequently $g(0)=0$. Let
\begin{align}
	g(z)=\alpha h(z)+\beta,&&z\geq0.
\end{align}
for some odd symmetric function $h(\cdot)$ and constants $\alpha,\beta$. Similarly, $g(z)=-\alpha h(-z)-\beta$, $z\leq 0$.
We can equivalently write the function as
\begin{align}
	g(z)=&\sign(z)(\alpha h(\abs{z})+\beta),
\end{align}
and its first derivative as
\begin{align}
	g'(z)=&\alpha h'(\abs{z}).
\end{align}
From \eqref{eq:dl}, for the convexity of $l_y(g(\cdot))$, we need the following to be nondecreasing in $z$:
\begin{align}
	(g(z)-y)g'(z)=&\bigg(\sign(z)(\alpha h(\abs{z})+\beta)-y\bigg)\alpha h'(\abs{z}),\\
	=&\alpha^2\sign(z) h(\abs{z})h'(\abs{z})
	\\&+\alpha h'(\abs{z})\bigg(\sign(z)\beta-y\bigg)
\end{align}
We simplify the expression with the following assumption.
\begin{assumption}
	Let $h(z)$ be an odd symmetric function such that
	\begin{align}
		h(\abs{z})h'(\abs{z})=\gamma,
	\end{align}
for all $z$ for some $\gamma$.
\end{assumption}
\noindent Hence, the following needs to be nondecreasing:
\begin{align}
	(g(z)-y)g'(z)=&\alpha^2\gamma\sign(z)+\alpha h'(\abs{z})\bigg(\sign(z)\beta-y\bigg)\label{eq:expr}
\end{align}
We observe that $\sign(z)$ is different but fixed when $z$ is positive or negative. Thus, the second term is of greater interest. 
\begin{assumption}\label{ass:h'a}
	Let $h(\cdot)$ be such that $h'(\abs{z})$ is nonincreasing with $\abs{z}$. 
\end{assumption}
If \autoref{ass:h'a} holds, $h'(z)$ will be nonincreasing when $z\in[0,\infty)$ and nondecreasing when $z\in(-\infty,0]$ because of odd symmetry. For the second term to be nondecreasing, we need $\beta-y\leq 0$ and $-\beta-y\geq 0$. Let us assume $-Y\leq y\leq Y$ for some constant $Y$. Then, setting $\beta=-Y$ takes care of the requirement.
Because of the continuity of $(g(z)-y)g'(z)$ at $z=0$, we have
\begin{align}
	-\alpha\gamma+h'(0)(Y-y)=&\alpha\gamma+h'(0)(-Y-y),
	\\-\alpha\gamma+h'(0)Y=&\alpha\gamma-h'(0)Y,
	\\h'(0)Y=&\alpha\gamma.
\end{align}
When this continuity condition is satisfied, the expression in \eqref{eq:expr} will be nondecreasing in $z$. 
\begin{remark}
	If a function $h(\cdot)$ satisfies the following conditions
	\begin{itemize}
		\item $h(z)=-h(-z), \forall z$
		\item $h(\abs{z})h'(\abs{z})=\gamma, \forall z$,
		\item $h'(\abs{z})$ is nonincreasing with $\abs{z}$,
		\item $h'(0)Y=\alpha\gamma$,
	\end{itemize}
	for some $\gamma,\alpha\in\mathbb{R}$ and $-Y\leq y\leq Y$; $l_y(\alpha h(z)-Y)$ is convex in $z$.
\end{remark}

One such function is given in the following result.
\begin{lemma}
	The nonlinear transform
	\begin{align}
		g_{\alpha,Y}(z)=\sign(z)(Y\sqrt{\alpha z+1}-Y)
	\end{align}
	makes $l_{y}(g_{\alpha,Y}(z))$ convex in $z$; when $-Y\leq y\leq Y$ and $\alpha>0$.
	\begin{proof}
		The function can be equivalently written as
		\begin{align}
			g_{\alpha,Y}(z)=\begin{cases}
				\displaystyle	Y{\sqrt{\alpha z+1}}-Y, &z\geq0\\
				\displaystyle	-Y{\sqrt{-\alpha z+1}}+Y, &z<0
			\end{cases}.
		\end{align}
		The first derivative is nonnegative and continuous, i.e.,
		\begin{align}
			g_{\alpha,Y}'(z)=\begin{cases}
				\displaystyle	\frac{Y\alpha}{2\sqrt{\alpha z+1}}, &z\geq0\\
				\displaystyle	\frac{Y\alpha}{2\sqrt{-\alpha z+1}}, &z<0
			\end{cases}.
		\end{align}
		Since $l_{y}'(g_{\alpha,Y}(z))=2(g_{\alpha,Y}(z)-y)g_{\alpha,Y}'(z)$, we will look at the monotonicity of $2(g_{\alpha,Y}(z)-y)g_{\alpha,Y}'(z)$, which is given by
		\begin{align}
			2(g_{\alpha,Y}(z)-y)g_{\alpha,Y}'(z)=\begin{cases}
				\displaystyle	\alpha Y^2-\frac{\alpha Y(Y+y)}{\sqrt{z+1}},&z\geq0\\
				\displaystyle	-\alpha Y^2+\frac{\alpha Y(Y-y)}{\sqrt{-z+1}},&z<0
			\end{cases}
		\end{align}
		Since $-Y\leq y\leq Y$, this is continuous and nondecreasing with $z$ since $-Y\leq y\leq Y$. Hence for the square loss $l_{y}(\cdot)$, the nonlinear transform $g_{\alpha,Y}(\cdot)$ preserves the convexity when $-Y\leq y\leq Y$ and $\alpha>0$.
	\end{proof}
\end{lemma}

\bibliographystyle{IEEEtran}
\bibliography{double_bib}

\end{document}